\documentclass{article}
\usepackage{microtype}
\usepackage{graphicx}

\usepackage{hyperref}

\usepackage[accepted]{icml2022}

\usepackage{amsmath}
\usepackage{amssymb}
\usepackage{mathtools}
\usepackage{amsthm}
\usepackage{mediabb}
\usepackage[capitalize,noabbrev]{cleveref}
\theoremstyle{plain}

\theoremstyle{definition}

\theoremstyle{remark}

\usepackage[textsize=tiny]{todonotes}

\usepackage{times}
\usepackage{soul}
\usepackage{url}

\usepackage[utf8]{inputenc}
\usepackage[small]{caption}
\usepackage{algorithm}
\usepackage{algorithmic}
\usepackage{here}
\usepackage{subfigmat}
\usepackage{bmpsize}

\usepackage{svg}
\usepackage{float}
\usepackage{natbib}

\usepackage{color}
\usepackage{tikz}
\usepackage{url}
\usepackage{mathtools}
\usepackage{here}
\usepackage{dcolumn}
\usepackage{autobreak}
\usetikzlibrary{bayesnet, arrows, decorations.markings,shapes.geometric, positioning}

\usepackage[T1]{fontenc}
\date{}
\newtheorem{theo}{Theorem}

\newcommand{\tbref}[1]{Table~\ref{#1}}

\newcommand{\figref}[1]{Figure~\ref{#1}}

\newcommand{\theoref}[1]{Theorem~\ref{#1}}

\newcommand{\algref}[1]{algorithm~\ref{#1}}

\renewcommand{\P}{\mathbb{P}}
\newcommand{\E}{\mathbb{E}}
\newcommand{\T}{\mathcal{T}}

\newcommand{\Sset}{\mathcal{S}}

\newcommand{\Aset}{\mathcal{A}}

\newcommand{\calW}{\mathcal{W}}

\newcommand{\secref}[1]{Section~\ref{#1}}

\tikzstyle{myvec}=[line width=1mm,draw=gray,-triangle 45,postaction={draw, line width=3mm, shorten >=4mm, -}]

\newcommand{\edit}[1]{#1}

\newcommand{\comm}[1]{}

\newcommand{\tuple}[1]{$( {#1} )$}

\graphicspath{
{./figure/}
}

\icmltitlerunning{Unsupervised Discovery of Continuous Skills on a Sphere}
\begin{document}
\twocolumn[

\icmltitle{Unsupervised Discovery of Continuous Skills on a Sphere}
\icmlsetsymbol{equal}{*}

\begin{icmlauthorlist}
\icmlauthor{Takahisa Imagawa}{mazda}
\icmlauthor{Takuya Hiraoka}{aist,nec}
\icmlauthor{Yoshimasa Tsuruoka}{aist,ut}
\end{icmlauthorlist}

\icmlaffiliation{mazda}{Mazda Motor Corporation, Hiroshima, Japan (This work was done when the author was at NEC-AIST collaboration laboratory)}
\icmlaffiliation{aist}{National Institute of Advanced Industrial Science and Technology, Tokyo, Japan}
\icmlaffiliation{nec}{NEC Central Research Laboratories, Kanagawa, Japan}
\icmlaffiliation{ut}{The University of Tokyo, Tokyo, Japan}

\icmlcorrespondingauthor{Takahisa Imagawa}{imagawa.takahisa@gmail.com}
\icmlkeywords{Reinforcement Learning, Unsupervised Learning, Self-Supervised Learning, Mutual Infomation}

\vskip 0.3in
]
\printAffiliationsAndNotice{}  

\graphicspath{
{./figures/}
}
\newcommand{\ourmet}{DISCS} 
\newcommand{\numocc}{the number of occupied cells}
\newcommand{\Numocc}{The number of occupied cells}

\begin{abstract}
  Recently, methods for learning diverse skills to generate various behaviors without external rewards have been actively studied as a form of unsupervised reinforcement learning.
  However, most of the existing methods learn a finite number of discrete skills, and thus the variety of behaviors that can be exhibited with the learned skills is limited.
  In this paper, we propose a novel method for learning potentially an infinite number of different skills, which is named \textit{discovery of continuous skills on a sphere} (DISCS).
  In DISCS, skills are learned by maximizing mutual information between skills and states, and each skill corresponds to a continuous value on a sphere.
  Because the representations of skills in DISCS are continuous, infinitely diverse skills could be learned.  
  We examine existing methods and DISCS in the MuJoCo Ant robot control environments and show that DISCS can learn much more diverse skills than the other methods.
\end{abstract}

\section{Introduction}\label{sec:intro}
\begin{figure}[tb]
  \centering
   \includegraphics[keepaspectratio,width=0.32\hsize]{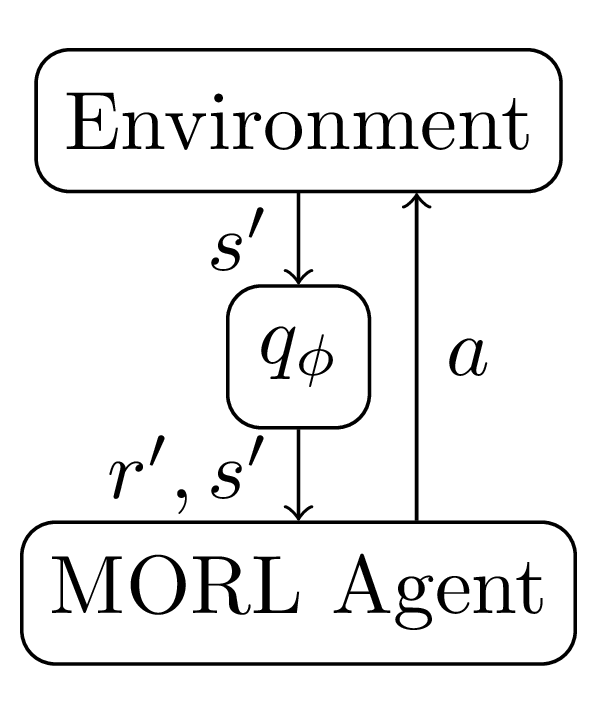}
\caption[]{\edit{Overview of our proposed method. Our agent learns policies by a multi-objective reinforcement learning method introduced in \secref{sec:mosac} and its reward vectors are generated by a method explained in \secref{sec:rewgen}.}}
  \label{fig:sketch}
\end{figure}
Deep reinforcement learning (RL) has shown promising results in various domains such as robotic control~\cite{openai2019solving,chen2021a} and games~\cite{silver2017mastering, berner2019dota, oriol2019alphastar}.
However, a typical RL agent learns each task from scratch by using external rewards in terms of how well the task is solved.
This is in stark contrast to the way humans explore the environment and learn various skills and strategies without such external evaluation.

To fill the gap, methods for learning various skills (i.e., potentially useful sequences of actions) without external rewards have been studied as a form of unsupervised or self-supervised RL.
These methods are important in practice since it is generally difficult and costly to design appropriate rewards for individual tasks and hence external rewards are not always available~\cite{roijers2013survey,dulac2019challenges}.
In such cases, task-agnostic skills learned by unsupervised RL help the agent to quickly solve the task once the the external rewards are provided.
Such skills are also useful in helping the agent to efficiently explore the environment.

A common approach to learning skills by unsupervised RL is to maximize mutual information between skills and states.
The existing mutual information-based methods differ in how they treat skills.
Specifically, a skill is treated as a discrete variable~\cite{gregor2016variational, eysenbach2018diversity, baumli2020relative, sharma2019dynamics},
a variable in the goal space~\cite{warde-farley2018unsupervised}, or a variable in a space in which state sequences are embedded by a variational autoencoder~\cite{campos2020explore, kim2021unsupervised}.
These methods have been shown to help the agent learn useful skills. 

\edit{Among the mutual information-based methods, a recently proposed one, Variational Intrinsic Successor FeatuRes (VISR)~\cite{hansen2020fast} has two advantages.  First, skills in VISR are continuous skills.  Since its skills are continuous, the agent can pontentially learn a myriad of skills.  Second, skills are learned in association with rewards (more specifically, as weights of the reward vectors).  By learning skills in a such manner, when external rewards are given, estimating the weights (i.e., appropriate skills for the given rewards) is relatively easy~\cite{barreto2018transfer,yang2019generalized}. }

While VISR has these advantages, it also has drawbacks.
VISR has been tested only in discrete action domains in the original and subsequent research~\cite{liu2021aps},
and according to the experimental results of \citet{kim2021unsupervised} in continuous action control environments, the diversity of skills learned by VISR was limited.
Futhermore, the analysis of unsupervised learning process itself (e.g., sample efficiency) has been rarely performed. 
Due to the computational cost of unsupervised learning, its sample efficiency is important, and so is the analysis from that perspective.

In this paper, we propose a new unsupervised RL method, \textit{discovery of continuous skills on a sphere} (\ourmet{}), which learns continuous skills as weights of reward vectors like VISR.
\edit{We show an overview of our method in \figref{fig:sketch}.}
We investigate the process of unsupervised learning in existing methods and \ourmet{} in the MuJoCo Ant robot control environment with continuous actions, and show that \ourmet{} can sample-efficiently learn various skills compared to existing methods.
We also show that learning skills in VISR is more difficult than \ourmet{} because of its generation of rewards.
Furthermore, we show that an existing discrete skill learning method with many skills cannot be a substitute for \ourmet{}.
In addition, we propose \textit{hindsight preference posterior sampling} (HIPPS) as one of the techniques of \ourmet{} and show that it helps learning in \ourmet{}.

The paper is organized as follows.
We introduce the background of \ourmet{}, multi-objective RL in \secref{sec:background} and details of \ourmet{} in \secref{sec:method}.
In \secref{sec:related}, related work including VISR and differences between VISR and \ourmet{} are introduced.
In \secref{sec:experiments}, experimental analysis and comparisons between existing methods and \ourmet{} are shown.
In \secref{sec:conclusion}, concluding remarks are given.

\section{Background}\label{sec:background}
This section briefly introduces multi-objective RL (MORL), upon which our method is based.

The tasks in MORL are modeled as multi-objective Markov decision processes (MOMDPs)~\cite{roijers2013survey}.
An MOMDP is an extension of well-known Markov decision processes (MDP)~\cite{sutton2018reinforcement}.
An MOMDP can be represented by a tuple \tuple{\Sset, \Aset, R, T, s_0, \calW, f_\calW}, where
$\Sset$ and $\Aset$ are the spaces of states and actions, respectively,
$s_0 \in \Sset$ is the initial state,
$r: \Sset \times \Aset \rightarrow \mathbb{R}^m$ is a reward vector function whose output dimension is $m$,
$T: \Sset \times \Aset \times \Sset$ is a function that determines the probability of transition to a state when an action is taken at a state, 
$\calW \subset \mathbb{R}^m$ is the space of preferences, and
$f_\calW: \calW \times \mathbb{R}^m \rightarrow \mathbb{R}$ is a scalarization function that transforms the total reward to a scalarized total reward according to a preference.
We consider the class of MOMDPs with a linear scalarization function.
That is, $f_\calW(w, {V}^\pi(s,w)) = w^T{V}^\pi(s,w)$, where ${V}^\pi(s,w)$ is $\E_{\pi_w}\left[ \sum_{t = 0} \gamma^t r(s_t,a_t) | s_0 = s, w\right]$, $w \in \calW$ and $\pi_w$ is a policy (a distribution over actions) for $w$.

The goal of MORL is to learn a policy that maximizes the total scalarized reward for each preference.
An MOMDP with only one preference corresponds to one MDP.
MORL is a framework for improving learning efficiency by learning the optimal policy set for the given preference set, rather than learning policies from scratch in individual MDPs.

In this paper, as in previous work~\cite{abels2019dynamic, yang2019generalized, chen2020two},  we focus on learning a preference conditional policy and Q-function, which returns the expected cumulative reward vectors for the policy.  
Also, we put constraints on $\calW$ to remove redundancy of $w$ (e.g., a multiplication of rewards leads to the same optimal policy).
For example, \citet{yang2019generalized} regularize the L1 norm of preferences.   
In our method, we regularize the L2 norm of preferences instead because of the tractability of distributions on $\calW$ and $\calW = \{w\: |\: ||w||_2 = 1, w \in \mathbb{R}^m \}$.

Note that an MORL agent learns preference conditional policies, which means that preference $w$ controls sequential actions (often referred to as a skill). 
Thus, we refer to $w$ as not only a preference but also a skill.

\section{Discovery of Continuous Skills on a Sphere}\label{sec:method}
In this section, we introduce three main components in \ourmet{}: 1) multi-objective soft actor-critic (MOSAC), 2) reward vector generation, and 3) their effective training, HIPPS.
\ourmet{} learns a policy by MOSAC, which is a simple extension of soft actor-critic (SAC)~\cite{haarnoja2018soft2} to MORL, which is one of the most sample-efficient off-policy RL methods.
A \ourmet{} agent learns how to generate reward vectors on the basis of mutual information between states and skills on a unit sphere, and the generated reward vectors are used for the learning in MOSAC.  \ourmet{} uses HIPPS, which aims to improve the sample efficiency of \ourmet{} by adding data sampled from the distribution (posterior) learned in the reward generation.
Pseudo code of \ourmet{} is shown in \secref{sec:implementation}.

\subsection{Multi-Objective Soft Actor-Critic}\label{sec:mosac}
\edit{An MOSAC agent collects data from the environment (rollouts), preserves them in a replay buffer.}
\edit{By using data in the replay buffer, the agent iteratively learns preference conditional policies and Q-functions as MORL.} 
The agent maximizes the sum of the policy entropy and the total reward, as in SAC.

For simplicity, we introduce $m+1$-dimensional extended reward vector and preference, whose the $0$-th dimension is reserved for the entropy of policy.
Let $\tilde{r}$ denote $(c, r_1, \dots, r_m)^\top$, where $c$ is typically $0$ and $r_i$ ($1 \leq i \leq m$) is the $i$-th element of the original reward vector, 
and $\tilde{w}$ denote $(1, w_1, \dots, w_m)^\top$, where $w_i$ ($1 \leq i \leq m$) is the $i$-th element of the original preference. 
Let $h^{\pi}(s',a',w)$ denote a vector $(-\alpha \log \pi(a'| s', w), 0, \dots, 0)^\top$ for entropy of its policy whose dimension is $m+1$, where $\alpha$ is the coefficient of the entropy.
The Q-function is updated based on a Bellman operation with reward and entropy vectors,
\begin{align}
  &{\T} Q^{\pi}(s, a,w) = \tilde{r}(s,a) + \gamma \E_{s'}[V^{\pi}(s',w)]\\
  &V^{\pi}(s',w) = \E_{\pi(a'|s',w)} [Q^{\pi}(s', a',w) + h^{\pi}(s', a',w)].
\end{align}
Applying Bellman operation $\T$ defined above repetitively leads to a fixed point because ${\T}$ is a contraction mapping (see e.g., ~\cite{bertsekas2012dynamic2}).  
In MOSAC, its policy is updated as follows:
\begin{align}
  \arg\min_{\pi' \in \Pi}\mathrm{D}_{\mathrm{KL}}\left(\pi'(a| s,  w) \middle\| \frac{\exp(\frac{\tilde{w}^\top}{\alpha} {Q}^{\pi}(s,a,w))}{Z^\pi(s,w)}\right) \label{pi_target}
\end{align}

For these updates, extensions of two theorems in SAC~\cite{haarnoja2018soft2} can be derived in the same way as the proofs in SAC from the fact that one $w$ corresponds to one MDP.
\begin{theo}
  For any $s \in \Sset, a \in \Aset, w \in W$ and $\pi$, 
  $\pi'$ which is updated by \eqref{pi_target}, then $\tilde{w}^\top(Q^{\pi'}(s,a,w) - Q^{\pi}(s,a,w)) \geq 0$, assuming $|\Aset| < \infty$.  
\label{theo:policy_improve}
\end{theo}
This means that $\pi$ can be improved by \eqref{pi_target}.
\begin{theo}
  Repeated application of the updates of Q-functions and policies converges to a policy $\pi^{\ast}$ such that $\tilde{w}^\top({Q}^{\pi^\ast}(s,a,w) - {Q}^{\pi}(s,a,w)) \geq 0$ for all $\pi$ and $s,a,w$, assuming $|\Aset| < \infty$. 
  \label{theo:iterative_convergence}
\end{theo}
In this paper, the above policy and Q-function are approximated by neural networks.
Let $Q_{\theta_Q}(s,a,w)$ denote a Q-function and $\pi_{\theta_\pi}(a|s,w)$ a policy, whose parameter vectors are $\theta_Q$ and $\theta_\pi$, respectively.

In the same way as SAC, as the target for a Q-function update, we use a Q-function with parameter $\bar{\theta}$.
$\bar{\theta}$ is an exponential moving average of $\theta_Q$ and updated as $\bar{\theta} \leftarrow \tau\theta_Q + (1- \tau)\bar{\theta}$.

The policy and Q-fuction are updated by minimizing losses, $\mathcal{L}_{\mbox{actor}}$ and $\mathcal{L}_{\mbox{critic}}$, which are respectively,
\begin{align}
   &\E\left[ \alpha \log \pi_{\theta_\pi}(a_t|s_t, w) - \tilde{w}^\top Q_{\theta_Q}(s_t,a_t,w)  \right], \mbox{and }
  \label{eq:actor_loss}\\
   &\E\left[ -(\tilde{w}^\top(Q_{\theta_Q}(s_t,a_t, w) - \hat{\T} Q_{\bar{\theta}}(s_t,a_t, w)))^2 \right],
  \label{eq:critic_loss}
\end{align}  
where $\E$ in the above equations mean the expectations over tuples, ${(w,s_t, a_t, s_{t+1})}$, which are sampled from the replay buffer,
and $\hat{\T} Q_{\bar{\theta}}(s_t,a_t, w)$ is
\begin{align}
     \tilde{r}(s_t, a_t) + \gamma \E_{a \sim \pi_w}[Q_{\bar{\theta}}(s_{t+1},a, w)+h^\pi(s_{t+1}, a, w)].
\end{align}
\comm{
}

\subsection{Reward Vector Generation by Mutual Information}\label{sec:rewgen}
In \ourmet{}, the agent learns diverse skills by maximizing $I(S_t, W)$ the mutual information between states and preferences.
Due to the use of MOSAC, the agent also aims to maximize the policy entropy, $\mathcal{H}(A_t|S_t, W)$.
Intuitively, maximizing $I(S_t, W)$ means to go to the preference's own state as much as possible.
$I(S_t;W)$ can be expressed as $\E[\log p(w | s_t) - \log p(w)]$.
We fix $p(w)$ as the uniform distribution, so $\log p(w)$ is constant and can be ignored.
Hence, our objective is maximizing the expected sum of $\log p(w | s_t) - \log \pi(a |s_t, w)$ and this value corresponds to a scalarized reward including the policy entropy.

The expected sum of the scalarized rewards, including the expectation over the entire preferences, which is denoted as $\eta(\pi)$, has the following lower bound:
\begin{align}
  {\eta}(\pi) \geq \eta_\phi(\pi), 
\end{align}
where ${\eta}(\pi)$ and $\eta_\phi(\pi)$ are 
\begin{align}
  &\E_{w,\pi_w}\left[\sum_{t=0}\gamma^{t} \log p(w | s_t) - \alpha\log \pi(a_t |s_t, w) \right],\mbox{and}\\
  &\E_{w,\pi_w}\left[\sum_{t=0}\gamma^{t} \log q_\phi(w | s_t) - \alpha\log \pi(a_t |s_t, w) \right],
\end{align}
respectively, and $\phi$ is a parameter vector.
This inequality can be derived by an inequation of KL-divergence, $\mathrm{D}_{\mathrm{KL}}(p(w|s_t) || q_\phi(w|s_t)) \geq 0$.
Hereafter, $q_\phi(w | s)$ is referred to as a discriminator.

We aim to improve $\eta_\phi(\pi)$ instead of $\eta(\pi)$ by updating the policy and Q-function and the discriminator.
Let us assume $\phi'$ is a parameter vector updated from $\phi$ and the following inequality holds:
\begin{align}
\eta_{\phi'}(\pi) - \eta_{\phi}(\pi) = \E\left[\sum_{t=0}\gamma^{t}\Delta (\log q(w | s_t))\right] \geq 0,  \label{eq:disc_obj}
\end{align}
where $\Delta(\log q(w|s_t)) = \log q_{\phi'}(w|s_t) - \log q_{\phi}(w|s_t)$.
Under fixed $\phi'$, i.e., a fixed reward function, \theoref{theo:policy_improve} (and \theoref{theo:iterative_convergence}) hold.
Thus, updating the policy $\pi$ to $\pi'$ by \eqref{pi_target}, Q-values of any $(s,a,w)$ improve and the following inequalities are derived:
\begin{align}
  \eta_{\phi'}(\pi') \geq  \eta_{\phi'}(\pi) \geq \eta_{\phi}(\pi).
\end{align}
These mean that the $\eta$ value can be improved monotonically under condition \eqref{eq:disc_obj}.
Therefore, we update $\phi$ to improve $\eta_\phi(\pi)$. 

As for our discriminator, we use the von Mises-Fisher distribution (vMF), because our preferences are on a unit sphere (recall \secref{sec:background}) and vMF is a common probability distribution defined there.  
vMF has two parameters, $\mu$ and $\kappa$, which are the mean direction and concentration parameters, respectively and let vMF$(\mu, \kappa)$ denote vMF with the parameters.
More concretely, our discriminator is as follows:
\begin{align}
  q_\phi(w | s) = C_m(\kappa_{\phi_2}(s)) \exp(\kappa_{\phi_2}(s) w^\top \mu_{\phi_1}(s)),
\end{align}
where
$\kappa_{\phi_2}(s)$ is a scalar value, $\mu_{\phi_1}(s)$ is a $m$-dimensional vector,
$C_m(\kappa) = \frac{\kappa^{m/2 -1}}{(2  \boldsymbol{\pi})^{m/2} I_{m/2-1}(\kappa)}$ is a normalization constant,
$\boldsymbol{\pi}$ is the ratio of a circle's circumference to its diameter,
and $I_{m/2-1}(\kappa)$ is the modified Bessel function of the first kind at order $m/2 -1$.

Note that $\log C_m(\kappa)$ is partially differentiable with respect to $\kappa$ as follows: 
\begin{align}
    \frac{\partial}{\partial \kappa}\log C_m(\kappa) 
  = - \frac{I_{m/2}(\kappa)}{I_{m/2-1}(\kappa)}\label{eq:delkappa}
\end{align}  
The equation above can be derived by using $\frac{\partial}{\partial \kappa} I_{m/2-1}(\kappa) = \frac{m/2 - 1}{\kappa} I_{m/2-1}(\kappa) + I_{m/2}(\kappa)$.
This means that the gradients with respect to $\phi_2$ can be backpropagated via Equation~\eqref{eq:delkappa}.

Now that we have defined $q_\phi$ as above, $\log q_\phi (w|s) = \tilde{w}^\top \tilde{r}_\phi$, where $\tilde{r}_\phi$ is
\begin{align}
\kappa_{\phi_2}(s)\left(\frac{\log C_m(\kappa_{\phi_2}(s))}{\kappa_{\phi_2}(s)}, \mu_{1,\phi_1}(s), \dots, \mu_{m,\phi_1}(s) \right)^\top,
\end{align}
\edit{where $\mu_{i, \phi_1}$ is $i$-th element of $\mu_{\phi_1}$}
and we use $\tilde{r}_\phi$ as a reward vector for MOSAC.

We use the following value as the loss of discriminator, and update the parameter vector $\phi$ to minimize the loss:
\begin{align}
  \mathcal{L}_{\mbox{disc}} :=  - \E_{(w,s_t) \sim D}\left[\log q_\phi(w | s_t) \right],  
  \label{eq:disc_loss}
\end{align}
where $\E_{(w,s_t) \sim D}$ means the expectation for $(w,s_t)$ sampled from the replay buffer, $D$.
The loss defined above is different from that in theoretical analysis around inequation~\eqref{eq:disc_obj} in some respects.
As for the update of the discriminator, in the theoretical analysis, we considered a discounted objective with $\gamma$, but in the implementation, it was not included.
Also, in the theoretical analysis, we considered updating the discriminator for the state distribution defined by the most recent policy, but using only the recent data would reduce the amount of them.  
In the implementation, the discriminator is updated by sampling from the entire replay buffer.
\edit{Data in the replay buffer are collected by the previous policies which are generally different from the latest policy.}
More details of these differences are examined in \secref{sec:recent}.
\begin{figure}[h]
  \begin{subfigmatrix}{2}
 \subfigure{
   \includegraphics[keepaspectratio, width=0.48\hsize]{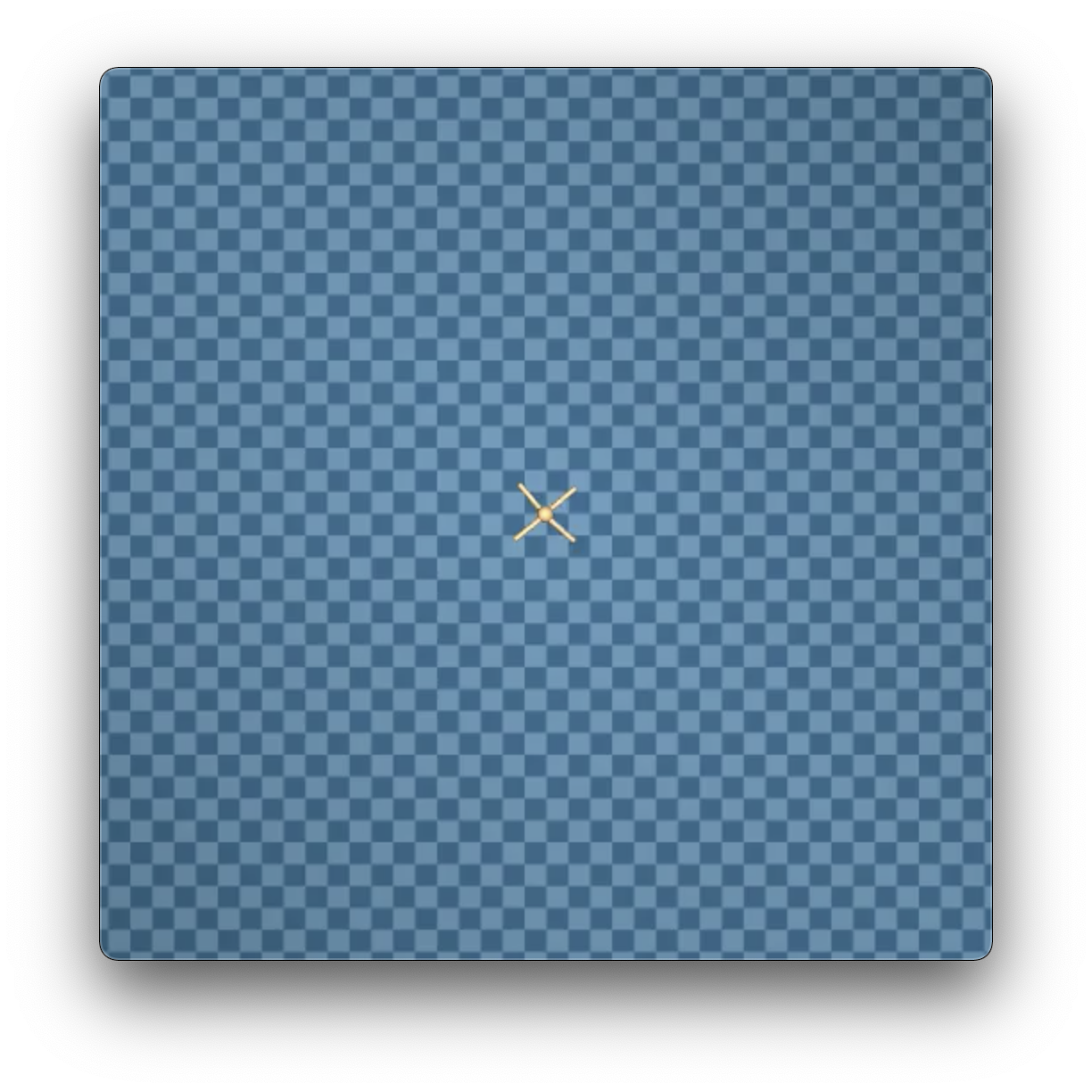}}
 \subfigure{
   \includegraphics[keepaspectratio, width=0.48\hsize]{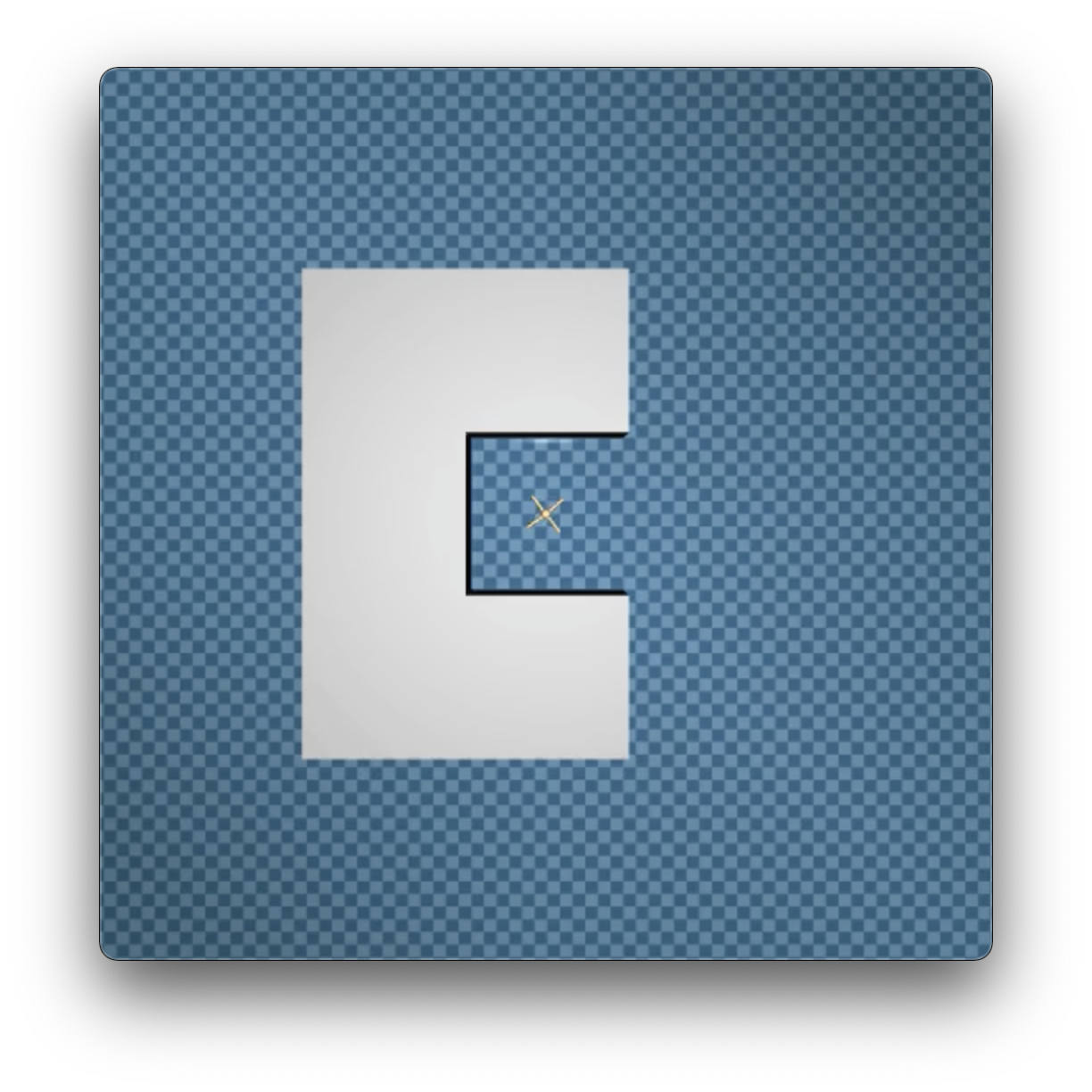}}
 \end{subfigmatrix}
\caption[]{Two types of environments, NoWall and U-Wall in our experiments.}
  \label{fig:domain}
\end{figure}

\begin{figure*}[t]
  \begin{subfigmatrix}{4}
 \subfigure[\numocc]{
   \includegraphics[width=0.24\hsize]{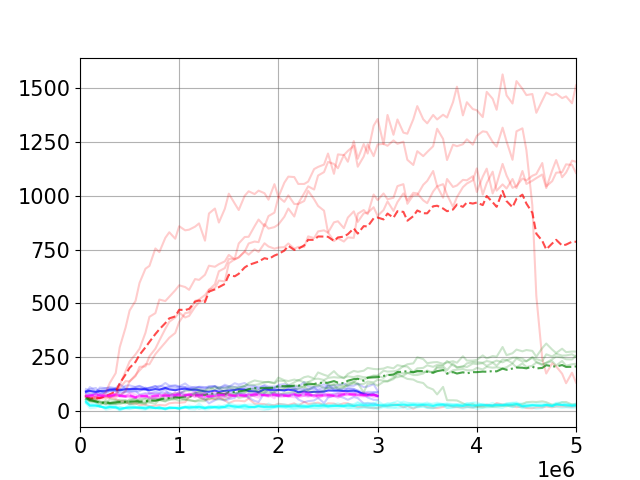}}
 \subfigure[discriminator loss]{
   \includegraphics[width=0.24\hsize]{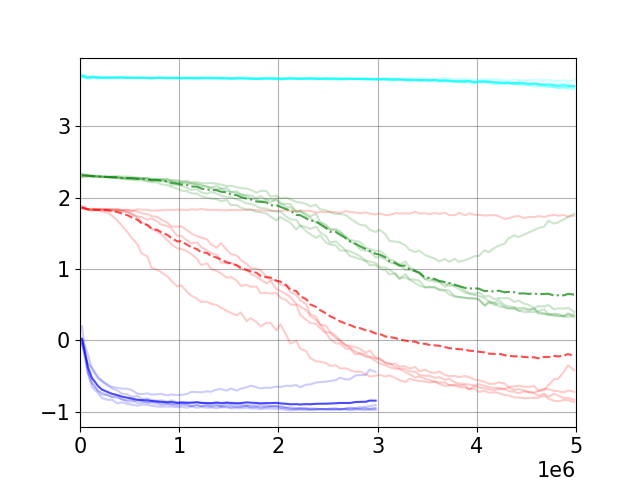}}
 \subfigure[average reward ]{
   \includegraphics[width=0.24\hsize]{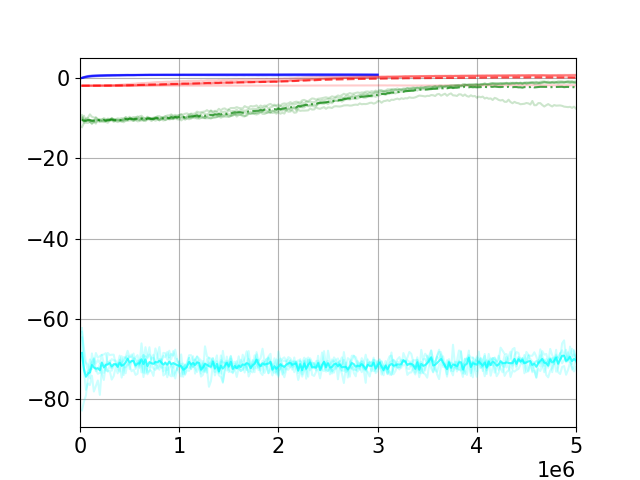}}
 \subfigure[ciritic loss ]{
   \includegraphics[width=0.24\hsize]{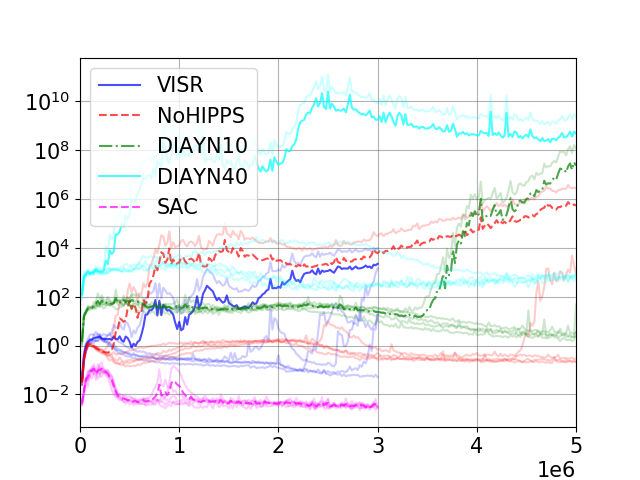}}
 \end{subfigmatrix}
\caption[]{Comparisons of learning curves in NoWall. Thin lines are actual data and thick lines are the averages of them.}
  \label{fig:comp_visr}
\end{figure*}
\begin{figure*}[t]
  \begin{subfigmatrix}{6}
 \subfigure[VISR]{
   \includegraphics[keepaspectratio]{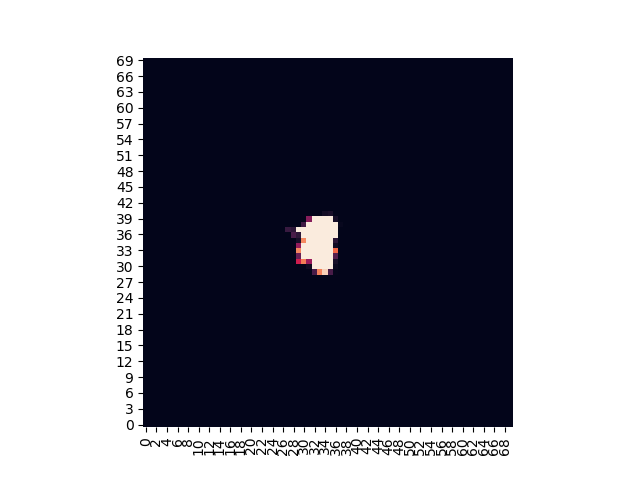}}
 \subfigure[NoHIPPS]{     
   \includegraphics[keepaspectratio]{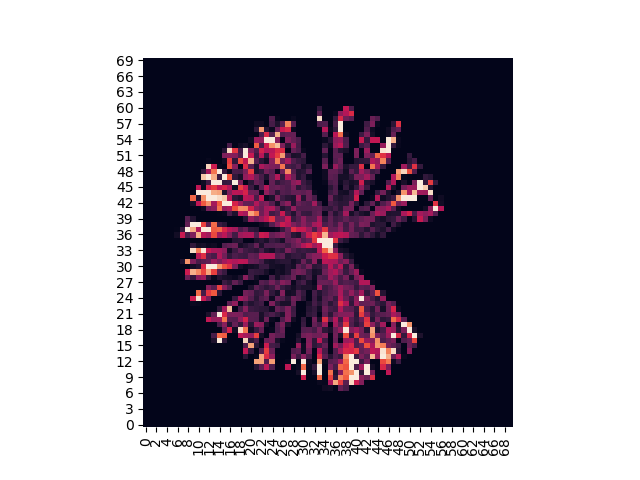}}
 \subfigure[HIPPS8]{      
   \includegraphics[keepaspectratio]{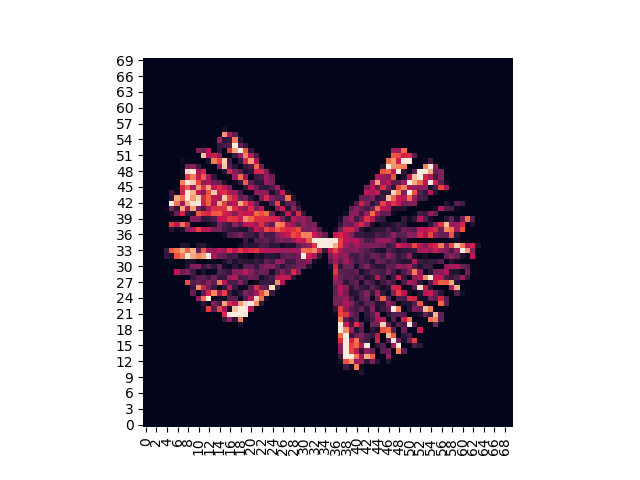}}
 \subfigure[DIAYN]{       
   \includegraphics[keepaspectratio]{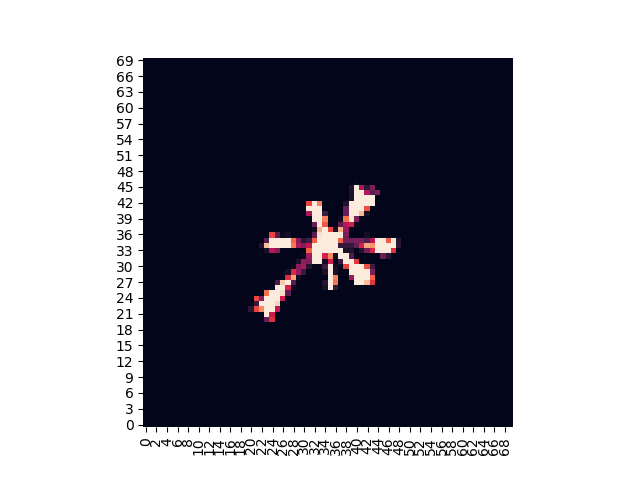}}
 \subfigure[DIAYN40]{     
   \includegraphics[keepaspectratio]{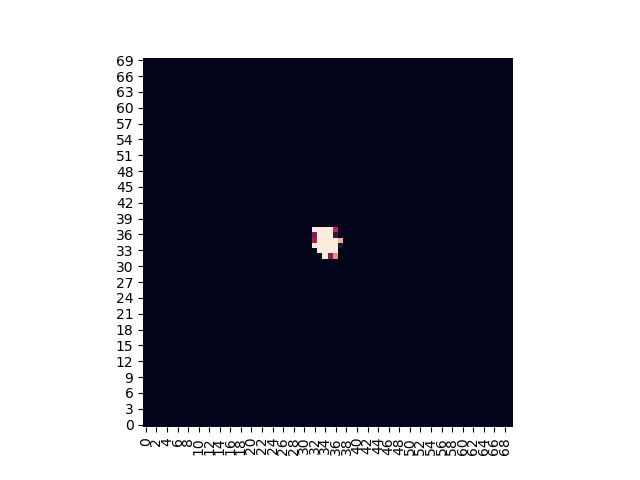}}
 \subfigure[SAC]{         
   \includegraphics[keepaspectratio]{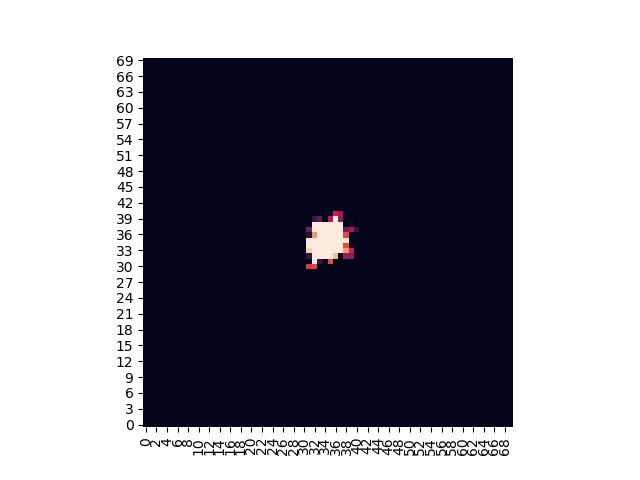}}
 \end{subfigmatrix}
\caption[]{Heatmaps in NoWall at 3 million timesteps in VISR, SAC and at 5 million timesteps in the other methods.} 
  \label{fig:comp_visr_heat}
\end{figure*}
\begin{figure*}[tb]
  \begin{subfigmatrix}{4}
 \subfigure[\numocc]{
   \includegraphics[width=0.24\hsize]{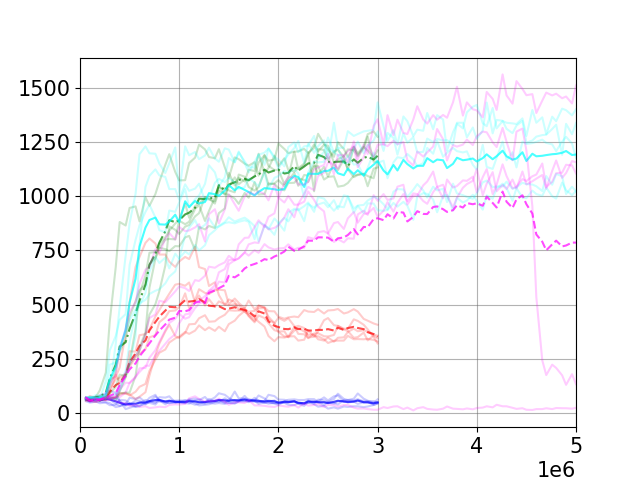}}
 \subfigure[discriminator loss]{
   \includegraphics[width=0.24\hsize]{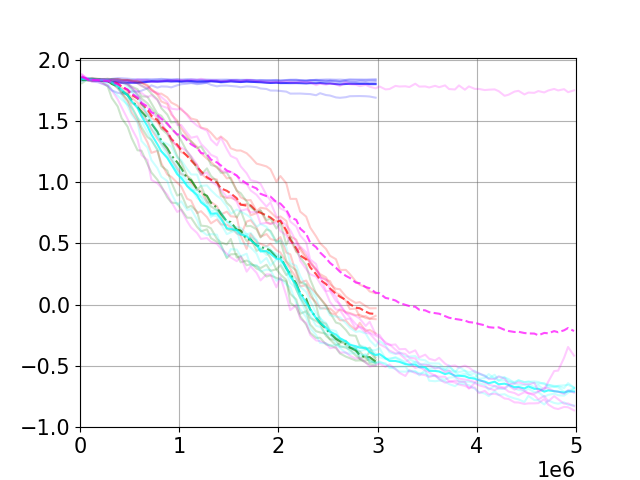}}
 \subfigure[average reward ]{
   \includegraphics[width=0.24\hsize]{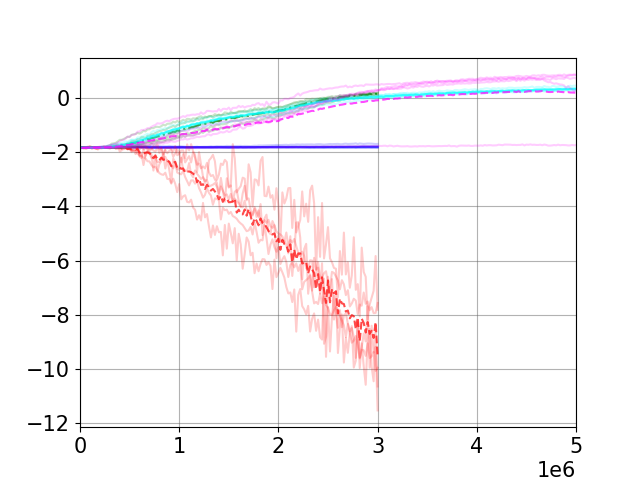}}
 \subfigure[ciritic loss ]{
   \includegraphics[width=0.24\hsize]{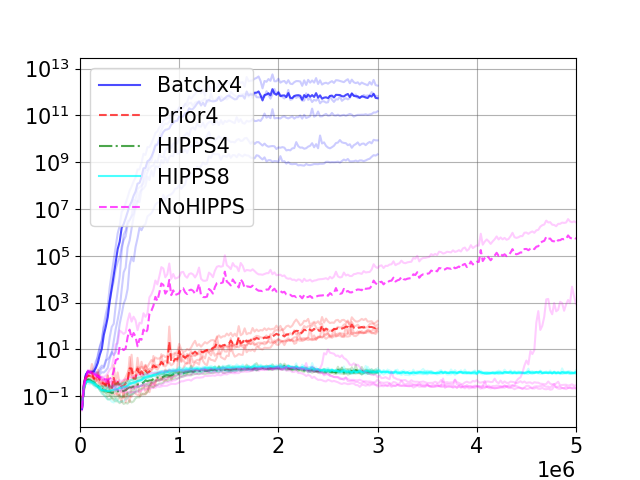}}
 \end{subfigmatrix}
\caption[]{Comparisons of learning curves in NoWall. Thin lines are actual data and thick lines are the averages of them.}
  \label{fig:comp_hipps}
\end{figure*}
\subsection{Hindsight Preference Posterior Sampling}\label{sec:hipps}
The discriminator can be optimized as described in the previous section.
However, learning (infinitely) many diverse policies may need much more data than learning a single policy.
We also propose a method to generate artificial data to learn policies effectively, named \textit{hindsight preference posterior sampling} (HIPPS).

Since our method is off-policy RL, it can learn from data that are not actually collected by the policy.
Therefore, it is possible to make learning more efficient by adding data.
HIPPS modifies the data in a hindsight manner, as in HER~\cite{andrychowicz2017hindsight}.
In HIPPS, in addition to actual data stored in the replay buffer $(w,s,a,s')$, additional data $(w',s,a,s')$, where $w'$ is a generated preference, are used for learning.

However, it may not be a good idea to train with arbitrary generated preferences.
\ourmet{} learns the policy, Q-function for each preference, as described in \secref{sec:mosac}.
However, learning to correctly approximate the Q-value for any $(w,s,a,s')$ is impractical and difficult\edit{, e.g.,} in terms of computational cost. 
Therefore, if we choose $w'$ poorly, the critic loss, for example, may be high for $(w',s,a,s')$.
As a result, the parameter update is affected by the loss, and the prediction of the Q-value for the actual data $(w,s,a,s')$ may become poor.
Thus, if we can choose a more plausible $w'$, our learning would be more efficient.
Motivated by this, we propose to sample additional preferences $w'$ from the discriminator, i.e., posterior, $q_\phi(w|s)$, and to use it as a tuple $(w',s,a,s')$ for the training of policy and Q-network.

We sample additional preferences from the projected normal distribution (PN)~\cite{mardia1975statistics, wang2013directional}, instead of sampling from vMF.
In general, sampling from vMF is difficult.
To address this difficulty, several sampling methods, including rejection sampling have been proposed~\cite{ulrich1984computer,kurz2015stochastic}.
We apply PN to HIPPS for its tractability.
PN is a probability distribution of $Y = \frac{X}{||X||_2}$, where $X$ is a random vector which follows the multivariate normal distribution $\mathcal{N}(\mu, \Sigma)$, which is denoted as PN($\mu, \Sigma$).
If $X$ is sampled from the multivariate normal distribution $\mathcal{N}(\mu, \frac{1}{\kappa}I)$, where $I$ is the identity matrix,
the distribution of $Y$ is vMF$(\mu, a \kappa)$ under condition, $||X||_2 = a$~\cite{mardia1975statistics}.
Moreover, PN and vMF converge to the uniform distribution and delta function as $\kappa$ approaches $0$ and $\infty$, respectively.
In addition to the above properties, the similarity of the two distributions is shown throughout experiments~\cite{campbell2019alignment}.
Thus we sample from PN($\mu, \frac{1}{\kappa}I$), instead of vMF. 
In \secref{sec:experiments}, we confirm that the approximation by PN is reasonable enough in terms of experimental results.

\section{Related Work}\label{sec:related}
Unsupervised RL methods based on mutual information are already introduced in \secref{sec:intro}. 
Among them, we will discuss the differences between the most related methods, VISR and DIAYN, and \ourmet{}.
Other related methods will also be briefly reviewed.

\textbf{VISR and DIAYN.}
VISR and DIAYN have the same objective, i.e., maximizing the mutual information between states and skills, as \ourmet{}.
While VISR is applied to Q-learning in the original paper~\cite{sutton2018reinforcement}, VISR is applied for MOSAC in this paper.
Apart from this difference, VISR can be seen as a special case of \ourmet{}, where $\kappa$ is $1$ in the discriminator and HIPPS is not applied.
In this case, $\log C_m(\kappa)$ is constant, so VISR ignores it.
By ignoring $\kappa$ and $\log C_m(\kappa)$, for the output of discriminator in VISR, $\log q_{\mathrm{VI}}(w|s)$, the following inequalities hold because of the L2 norm constraint: $-1 \leq \log q_{\mathrm{VI}}(w|s) = w^\top\mu(s) \leq 1$. 
To learn more fine-grained skills,
it is necessary to change the reward more finely according to the differences in the distribution of states induced by the skill, but this is difficult if $\kappa$ is constant, \edit{i.e., $\kappa$ in the distributions generated by the discriminator are the same value in any states}. 
DIAYN can also be seen as a special case of \ourmet{}, where its skill $z$ is a discrete variable and it does not deal with reward vectors.
Its discriminator's outcomes, $\log q_{\mathrm{DI}}(z|s_t)$, are used for its rewards. 
Also, HIPPS is not applied for DIAYN.

\textbf{Reward vectors.}
The existing methods for MORL
\cite{roijers2014bounded,mossalam2016multi,xu2020prediction,cao2021efficient}
and succussor features (SF)
\cite{barreto2017successor,borsa2018universal,hunt2019composing,barreto2019option,zahavy2021discovering}
are related in terms of using reward vectors.
In conventional SF settings, the agent optimizes its policy under condition that scalar rewards are given.
The SF agent approximates the reward by $w^\top \phi$, where $w$ and $\phi$ are a weight vector and a reward vector, respectively and learns the policy that maximizes total rewards.

\textbf{Other viewpoints.}
Our method is related to hierarchical RL~\cite{barto2003recent},
although our skill is only chosen at the initial state.
In addition, our method gradually changes rewards, which corresponds to gradually changing tasks. This is relevant to curriculum learning~\cite{narvekar2020curriculum}.
Our method is also relevant to intrinsic motivation and curiosity~\cite{schmidhuber2006developmental, bellemare2016unifying}, as the agent itself generates the reward.
As for vMF, \citet{kumar2018mises} applied it for natural language generation tasks.
As for the preference conditional Q-function in our method, the Q-functions in \citet{schaul2015universal} and \citet{borsa2018universal} are similar to ours, although their studies are not about unsupervised RL.

\section{Experiments}\label{sec:experiments}
In this section, we mainly examine the following questions:
1) Why is VISR difficult to learn diverse skills?
2) Can diverse skills be learned efficiently by the discrete skill learning method, DIAYN?  
3) \edit{How much can \ourmet{} outperform these methods?}
and 4) How can HIPPS help learning by \ourmet{}?   We also examine different update methods for discriminator in \ourmet{}.

We conducted experiments in the MuJoCo Ant robot control environments shown in \figref{fig:domain}.
In these environments, agents cannot get any rewards from them.
We ran five trials with different random seeds. 
In the experiments, to evaluate how diverse the learned skills is, we discretize the x-y positions of agents in rollouts and show heatmaps about the positions.
The episode length was set to 500 timesteps, and heatmaps were drawn for every 100 episodes, i.e., 0.05 million timesteps of data. 
In addition, to analyze the progress of the diverse skill learning, we measure the number of the discretized x-y positions whose visitation counts are positive (we refer to it as \numocc{}).
Also, we analyze the discrimination loss, critic loss and the average of scalarized rewards in the batch data excluding the policy entropy bonus.
In our experiments, all discriminators are trained with ``x-y prior'', which means that the inputs of the discriminators are x-y positions instead of states. 
In general, the state space is large, so it is difficult to learn skills without x-y prior that are diverse in terms of x-y positions.
In fact, in the experiments in DIAYN and DADS~\cite{eysenbach2018diversity, sharma2019dynamics}, the agents could not learn diverse skills in terms of x-y position without it.
\begin{figure*}[tb]
  \begin{subfigmatrix}{4}
 \subfigure[\numocc]{
   \includegraphics[width=0.24\hsize]{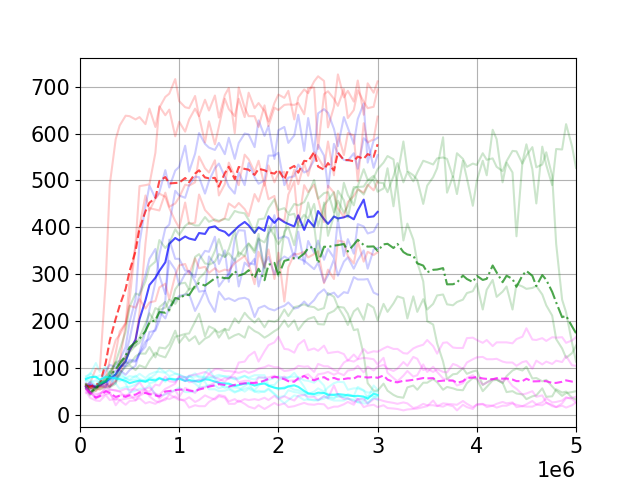}}
 \subfigure[discriminator loss]{
   \includegraphics[width=0.24\hsize]{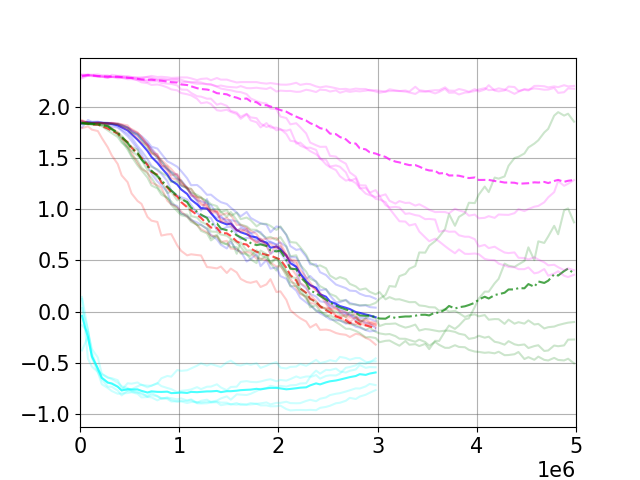}}
 \subfigure[average reward ]{
   \includegraphics[width=0.24\hsize]{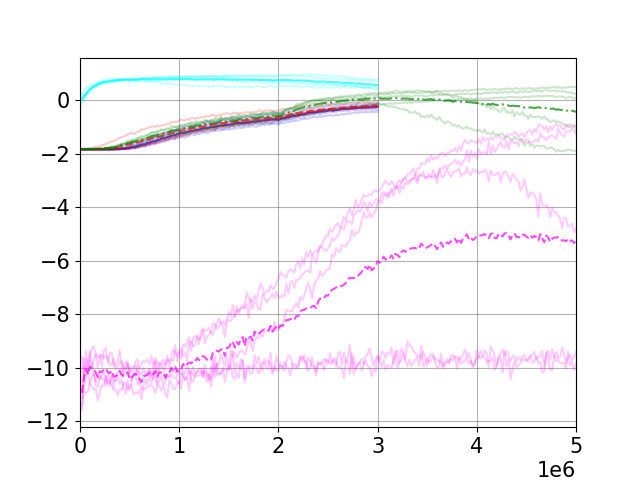}}
 \subfigure[ciritic loss ]{                 
   \includegraphics[width=0.24\hsize]{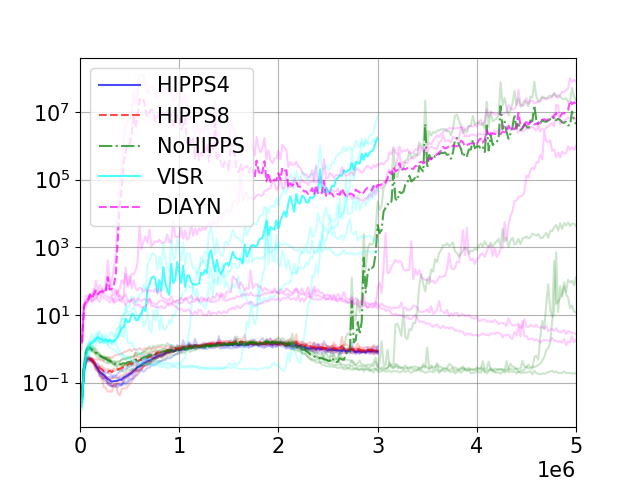}}
 \end{subfigmatrix}
\caption[]{Comparisons of learning curves in U-Wall. Thin lines are actual data and thick lines are the averages of them.}
  \label{fig:comp_uobs}
\end{figure*}
\begin{figure*}[tb]
  \begin{subfigmatrix}{5}
 \subfigure[VISR at 1M]{
   \includegraphics[keepaspectratio]{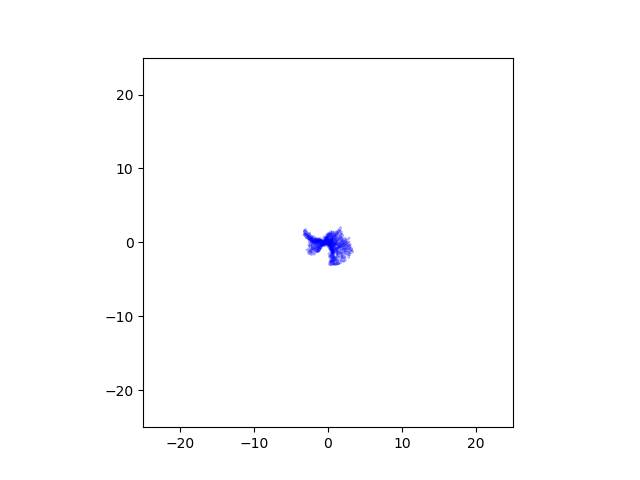}}
 \subfigure[VISR at 3M]{
   \includegraphics[keepaspectratio]{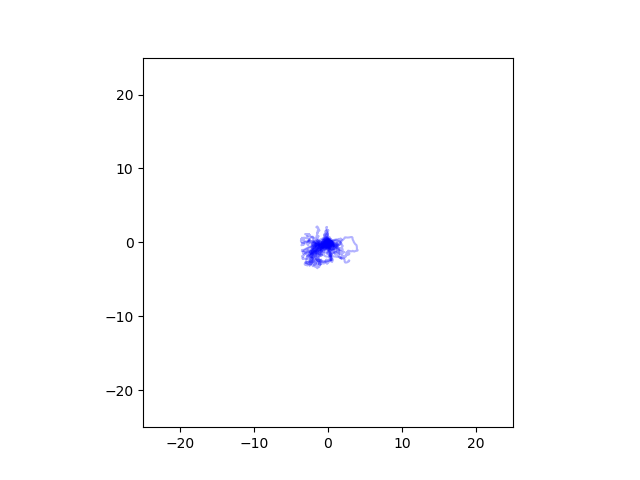}}
 \subfigure[NoHIPPS]{
   \includegraphics[keepaspectratio]{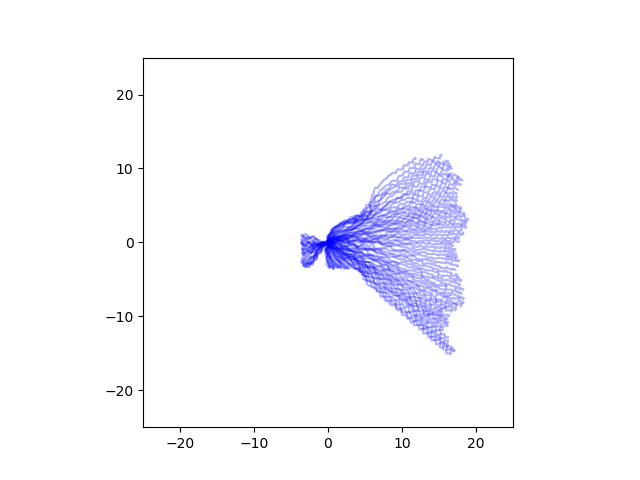}}
 \subfigure[HIPPS8]{
   \includegraphics[keepaspectratio]{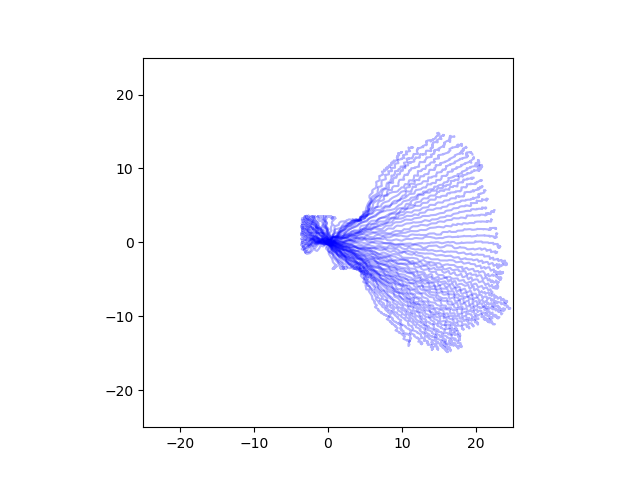}}
 \subfigure[DIAYN]{
   \includegraphics[keepaspectratio]{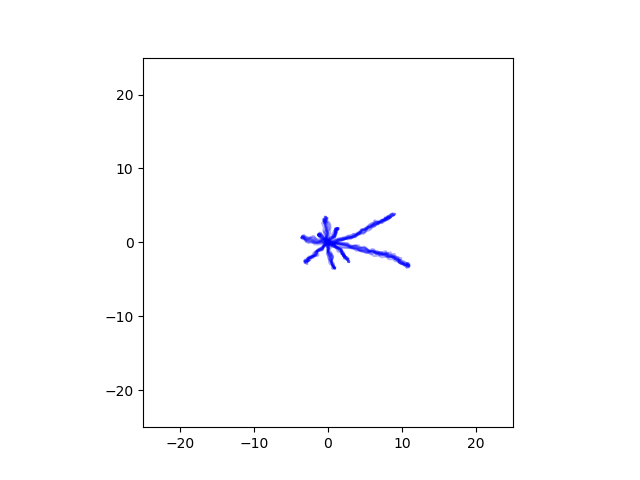}}
 \end{subfigmatrix}
\caption[]{Trajectories in U-Wall at 5 million timesteps in DIAYN and 3 million timesteps in HIPPS8, NoHIPPS, and VISR. Trajectories of VISR at 1 million timesteps are also shown.}
  \label{fig:comp_uobs_traj}
\end{figure*}

\begin{figure*}[tb]
  \begin{subfigmatrix}{4}
 \subfigure[\numocc]{
   \includegraphics[width=0.24\hsize]{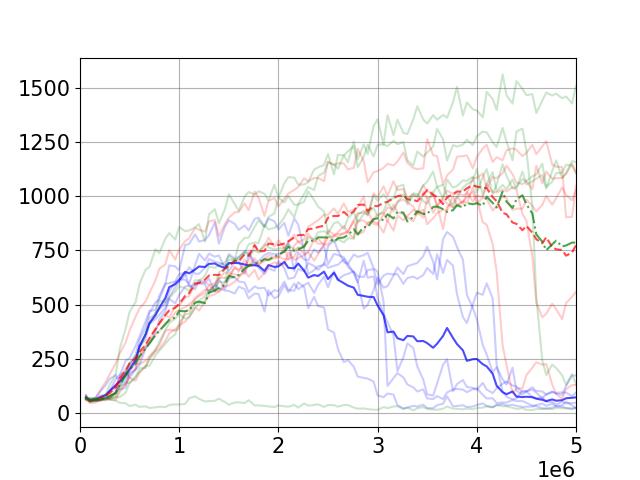}}
 \subfigure[discriminator loss]{
   \includegraphics[width=0.24\hsize]{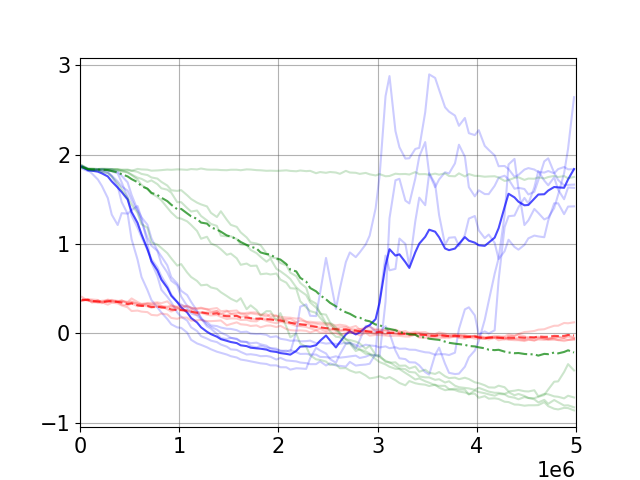}}
 \subfigure[average reward]{
   \includegraphics[width=0.24\hsize]{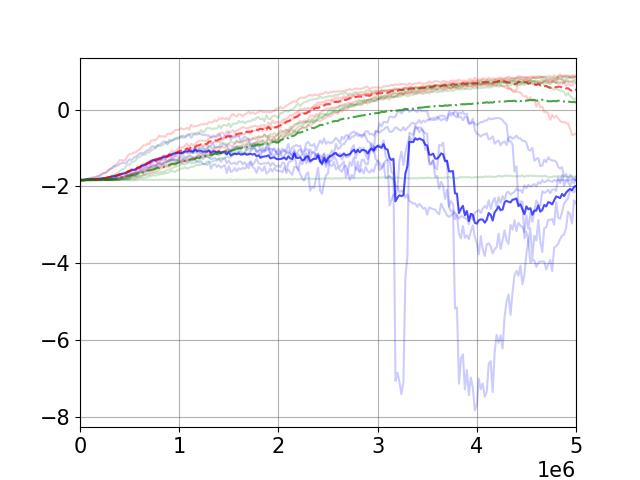}}
 \subfigure[ciritic loss]{                 
   \includegraphics[width=0.24\hsize]{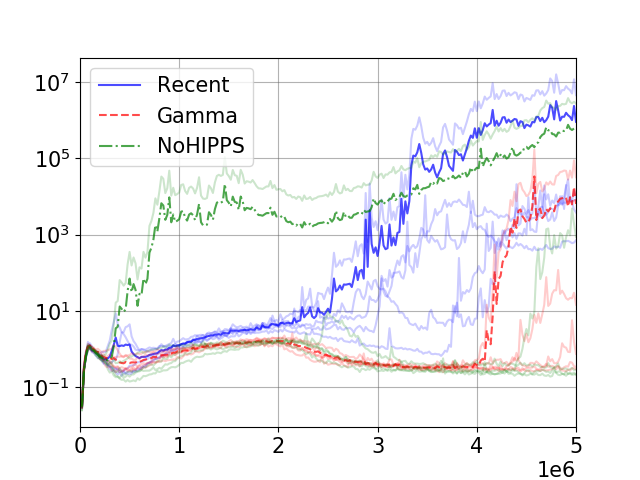}}
 \end{subfigmatrix}
\caption[]{Comparisons of learning curves in NoWall. Thin lines are actual data and thick lines are the averages of them.}
  \label{fig:comp_recent}
\end{figure*}

\subsection{Difficulties of VISR and DIAYN}\label{sec:difficulties}
We analyze why VISR does not work in the NoWall environment and examine whether a method for learning a large number of discrete skills, e.g., DIAYN, can be a substitute for that for learning continuous skills, e.g., \ourmet{}.
We compare VISR and DIAYN, in which the numbers of discrete skills are $10$ and $40$, with \ourmet{} without HIPPS (NoHIPPS) because of the similarities mentioned in \secref{sec:related}.
In addition, we show the performance of SAC, where there is no reward other than entropy of policy.  
The results are shown in \figref{fig:comp_visr}.

\Numocc{} of VISR was slightly larger than that of SAC.
Although one of the trials of NoHIPPS failed to learn, the other trails show that much more diverse skills were learned than VISR.
Also, the heatmaps (\figref{fig:comp_visr_heat}) showed that the area covered by VISR was much smaller than NoHIPPS. 
VISR was stable in terms of critic loss, except for the last 1 million timesteps. 
From this, it appears that the critic learning is fine.
We can see that the discrimination loss of VISR has decreased to around $-1$, which means the outputs of discriminator are nearly the minimum value (recall \secref{sec:related}).
What this means is that in order to learn more different skills, it is necessary to identify small differences in the rewards (i.e., the output of the discriminator) and learn a policy that reflects those differences.
In the same way, the discriminator also needs to reflect the differences in the state distribution defined by the policy for each skill. 
For those reasons, it is quite difficult to learn diverse skills by VISR.
On the other hand, the discrimination loss in NoHIPPS was decreasing and much larger than its minimum value which was $-\infty$ in theory ($-6 \log 10$ in our implementation).

\Numocc{} in DIAYN10 was much larger than in VISR but its sample efficiency was worse than NoHIPPS.
\Numocc{} of DIAYN40 was almost the same throughout the trials.
The discrimination loss of DIAYN40 finally began to decrease after around 5 million timesteps, which indicates that DIAYN40 was starting to learn diverse skills.
These results show that the learning in DIAYN needs more samples when the number of skills is increased, while that is not the case in \ourmet{}.

\subsection{HIPPS}\label{sec:hipps}
We examine whether \ourmet{} learns efficiently with more data by HIPPS in the NoWall environment.  
We also examine why the posterior is important in HIPPS by comparing it to the case where we sample from the prior rather than the posterior.
In addition, we compare \ourmet{} with HIPPS to NoHIPPS with a large batch and \edit{show that} simply increasing the batch size is not helpful.
Note that \ourmet{} with HIPPS uses a larger batch than NoHIPPS owing to its additional preferences. 
The results are shown in \figref{fig:comp_hipps}.
Batchx4 in the figure means simply quadrupling the batch size without using HIPPS.
HIPPS4/HIPPS8 means that HIPPS with 3/7 preferences are sampled for each tuple (and the total batch size is increased by 4/8 times).
Prior4 is a variant of HIPPS4 where the prior is used for the preference sampling instead of the posterior.

\Numocc{} of HIPPS4,8 and NoHIPPS were larger than the other methods.
In particular, HIPPS4,8 showed that the critic losses were low and \numocc{} were high in all trials.
The critic loss of Batchx4 was huge, which may be due to overtraining on the same data and one of the reasons for the failure in learning by Batchx4.  

\edit{In Prior4, \numocc{} started to decrease from about 1 million timesteps, and the critic loss started to increase} from about 0.5 million timesteps.
The discriminator loss of Prior4 decreased, which suggests that the state distribution changed with each preference and that the discriminator was able to correctly discriminate against the state distribution and that the distribution of discriminator became peaky.
On the other hand, the average reward of Prior4 decreased, which indicates that preferences with lower probability in terms of the distribution of discriminator were sampled.
These results support the claim made in \secref{sec:hipps} that sampling less plausible (in terms of the distribution of posterior) preferences from the prior increases the loss of critics and that it has negative effects on the learning.

\subsection{Comparisons in Environment with Obstacle}\label{sec:uobs}
When dealing with complex problems such as controlling ant robots in unsupervised RL, comparisons have been made mainly in tasks without obstacles.
In this work, we investigate how much the robot can bypass the obstacles by using U-Wall in \figref{fig:domain}.

We compare the results of the existing methods, DIAYN and VISR with \ourmet{}.
As confirmed in \edit{the results in \secref{sec:difficulties}}, DIAYN learns more slowly when the number of skills is increased. 
In this comparison, the number of skills in DIAYN was set to $10$.
The learning curves for NoHIPPS and DIAYN were measured up to 5 million timesteps, \edit{while those for other methods were 3 millon timesteps.}
The results are shown in \figref{fig:comp_uobs}.

\Numocc{} of \ourmet{} increased quickly.
In particular, \ourmet{} with HIPPS shows better results than the other methods while \numocc{} of NoHIPPS decreased from around 3 million timesteps and its critic loss increased.  
For DIAYN and VISR, the critic loss also increased during the skill learning process.
As for VISR, \numocc{} decreased after 1 million timesteps.
\edit{These results show that skill learning in U-wall tends to be unstable and difficult.}
On the other hand, the critic loss was stable in \ourmet{} with HIPPS.

For more detailed analysis, instead of heatmaps, we show trajectories when $100$ different generated skills were executed in \figref{fig:comp_uobs_traj} (heatmaps are shown in \secref{sec:addexp}).
The execution of skills was deterministic, i.e., the action was executed whose probability in the skill conditional policy was the highest. 
In the DIAYN case, $10$ different skills were executed $10$ times for each skill.
As for VISR, trajectories at 1 million timesteps, the timesteps before \numocc{} of VISR started to decrease, are also shown.  The trajectories of VISR at 1 million timesteps showed diverse behaviors although their covered area was limited.
As for the trajectories of DIAYN, the same skills showed almost the same trajectories and their diversity was limited. 
Compared with them, the results of \ourmet{} showed that it learned a variety of skills.

\subsection{Detailed Analysis of Discriminator Updates}\label{sec:recent}
We updated $\phi$ in a way that minimizes \eqref{eq:disc_loss}.
This deviates from theoretical analysis in \secref{sec:rewgen} in the following aspects.
1) All data in the replay buffer are sampled for the update.
2) It does not consider the discount of the reward by $\gamma$.
We examine these deviations.

With respect to the first point, we examine the performance when the data used to update the discriminator are limited to the latest data (Recent).
From the theoretical analysis in \secref{sec:rewgen}, it is ideal to update the discriminator with the latest policy data to increase its value, but on the other hand, the more we limit the data used to the latest one, the less data we can use.
As the latest data, we sampled from the recent 0.1 million steps data.

In addition, with respect to the second point, we examine a variant of the discriminator update where the rewards in the discrimination loss are discounted by $\gamma$ (Gamma).
Even if the deviation of the first point is ignored and assumed that the data are the latest, because the reward is not discounted by $\gamma$, an estimate of $\eta_\phi(\pi)$ is biased as discussed by ~\citet{thomas2014bias}.
\edit{To consider the discount of the reward}, we also keep timesteps $t$ in the replay buffer, and use $- \gamma^t \log q_\phi(w|s_t)$ as the loss for the sampled $(w, s_t, t)$. 
The results are shown in \figref{fig:comp_recent}.

The critic loss was more likely to increase when using only recent data than when using the entire replay buffer.
One possible explanation for \edit{these results is} catastrophic forgetting, where the learned \edit{relationships between inputs and outputs of the neural networks are} forgotten and cannot be reused, so the output of discriminator is not stable.
A method in \citet{abels2019dynamic} may alleviate the catastrophic forgetting, where mainly the latest data are used for the training, but also older data are used.
For simplicity, however, we sampled from the entire replay buffer to train the discriminator.

The performance about \numocc{} in Gamma was almost the same as that of NoHIPPS.
Although the estimation is biased, in our discriminator updates, we ignored the discount by $\gamma$ in \eqref{eq:disc_loss}, because it was also ignored in the discriminator loss in VISR and DIAYN and the performances of Gamma and NoHIPPS were almost the same.

\section{Conclusion}\label{sec:conclusion}
In this paper, we proposed \ourmet{}, an unsupervised RL method for learning skills, and HIPPS, a method for effective training in \ourmet{}.
\ourmet{} is different from most of the existing methods in that it has a clear correspondence with reward, it is a continuous skill learning method, and it uses HIPPS.
We conducted experiments in the MuJoCo Ant robot control environment with continuous actions and analyzed the process of unsupervised learning.
Through the analysis of the experiments, we showed that the existing method, VISR, has difficulty learning diverse skills due to the low expressive power of the discriminator, and that increasing the expressive power of the discriminator like \ourmet{} is important.
In addition, through the analysis of DIAYN, we showed that the learning became slower when the number of skills in DIAYN was increased.
This indicates that learning many discrete skills does not substitute for learning continuous skills.
Moreover, we examined \ourmet{} with and without HIPPS and showed that HIPPS contributed efficient and stable learning of skills in \ourmet{}.
\bibliographystyle{icml2022}
\bibliography{database}
\appendix
\onecolumn
\section{Additional Experiments}\label{sec:addexp}
In this section, we show additional experimental results which are omitted in the main article.
\subsection{\edit{Trajectories and Heatmaps}}
\begin{figure*}[tb]%
  \begin{subfigmatrix}{5}
 \subfigure[VISR]{
   \includegraphics[keepaspectratio]{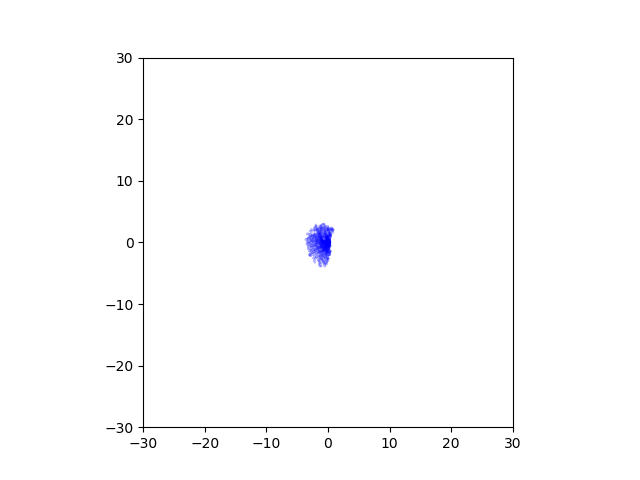}}
 \subfigure[NoHIPPS]{
   \includegraphics[keepaspectratio]{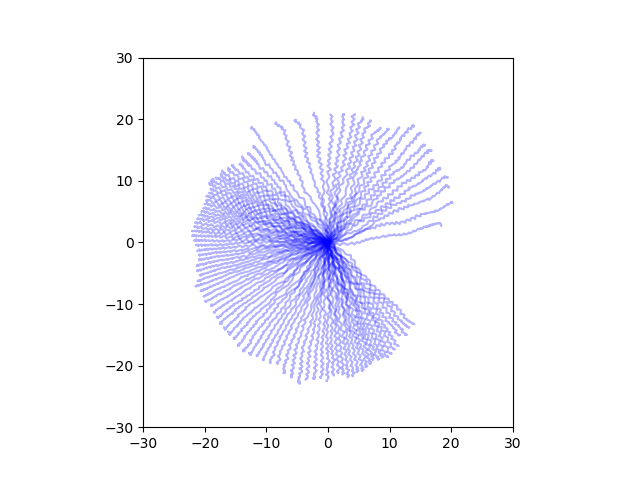}}
 \subfigure[HIPPS8]{
   \includegraphics[keepaspectratio]{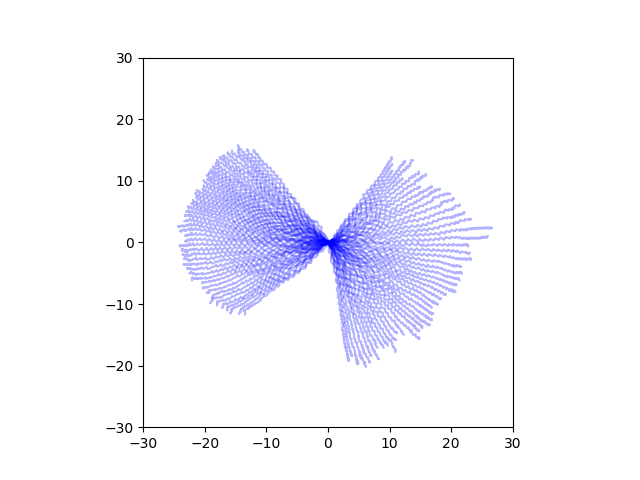}}
 \subfigure[DIAYN]{
   \includegraphics[keepaspectratio]{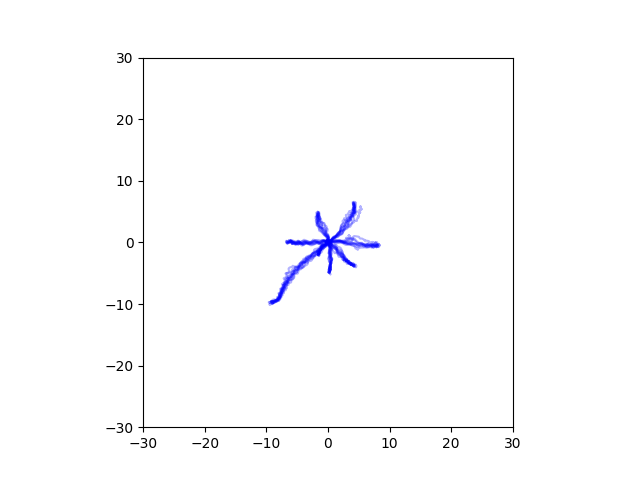}}
 \subfigure[DIAYN40]{
   \includegraphics[keepaspectratio]{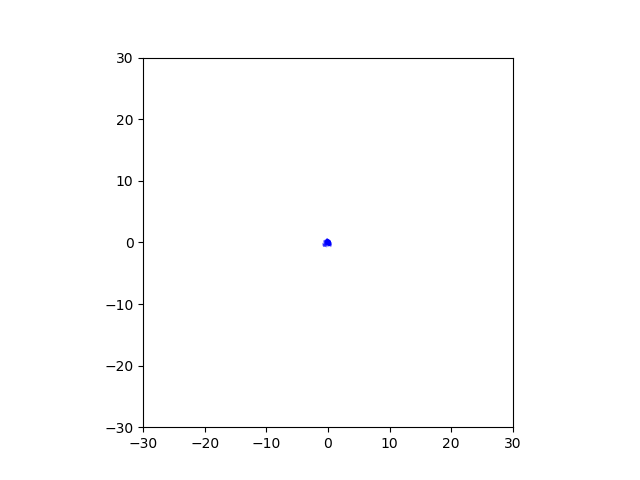}}
 \end{subfigmatrix}
 \caption[]{Trajectories in NoWall at 3 million timesteps in VISR and at 5 millon timesteps in the other methods.}
  \label{fig:comp_nowall_traj}
\end{figure*}

First, we show data about trajectories of learned skills in NoWall environment in \figref{fig:comp_nowall_traj}.
The trajectories are drawn in the same way as those in U-Wall environment (\figref{fig:comp_uobs_traj}).
The results are almost same as those of heatmaps (\figref{fig:comp_visr_heat}) except for DIAYN.
In particular, the result of DIAYN40 show that the agent cannot move at all in any direction.
Note that actions are chosen deterministically in evaluations for drawing trajectories as explained in \secref{sec:uobs}. 
The trajectories VISR cover a limited area, however they are more diverse than those of DIAYN40.

\begin{figure*}[tb]
  \begin{subfigmatrix}{5}
 \subfigure[VISR at 1M]{
   \includegraphics[keepaspectratio]{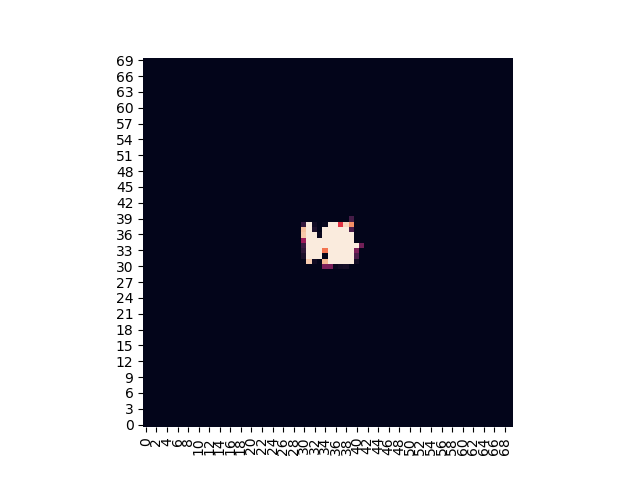}}
 \subfigure[VISR at 3M]{
   \includegraphics[keepaspectratio]{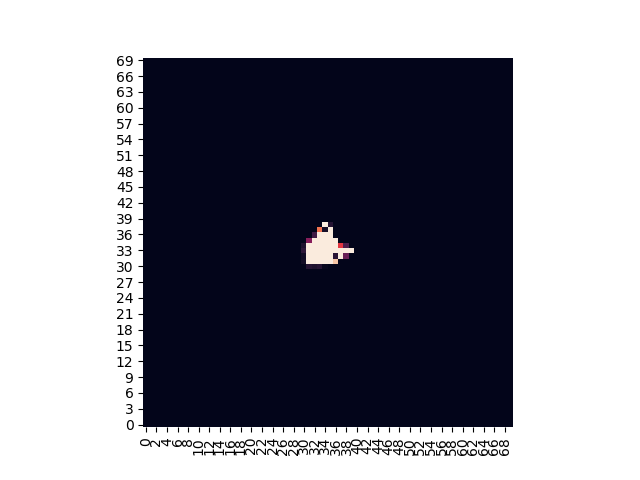}}
 \subfigure[NoHIPPS]{
   \includegraphics[keepaspectratio]{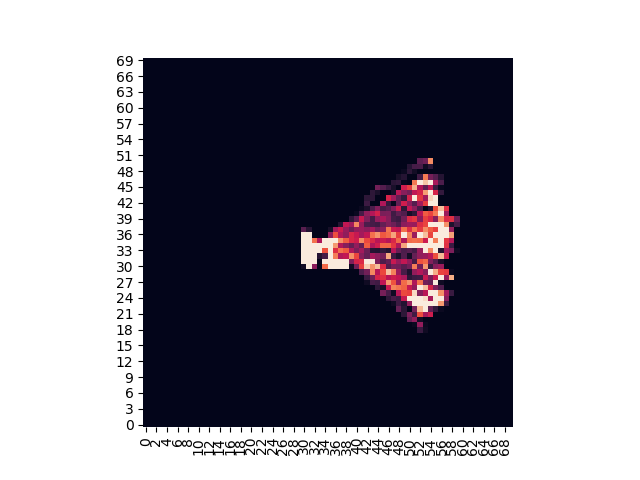}}
 \subfigure[HIPPS8]{
   \includegraphics[keepaspectratio]{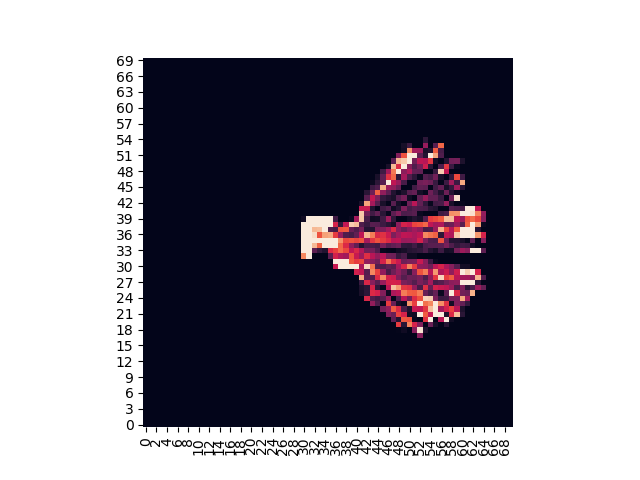}}
 \subfigure[DIAYN]{
   \includegraphics[keepaspectratio]{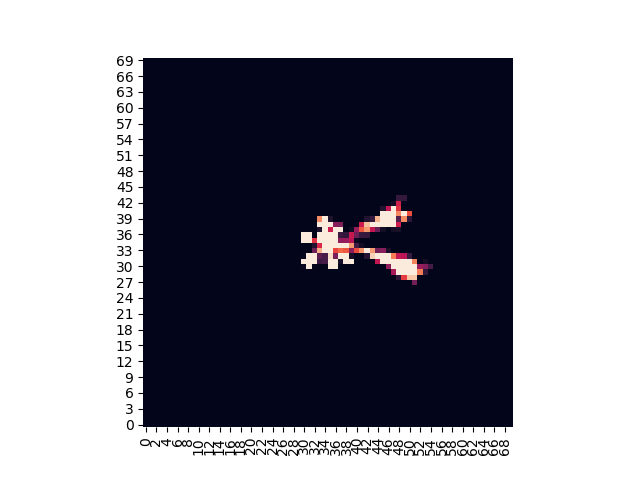}}
\end{subfigmatrix}
\caption[]{Heatmaps in U-Wall at 5 million timesteps in DIAYN and at 3 million timesteps in HIPPS8, NoHIPPS, and VISR. Trajectories of VISR at 1 million timesteps are also shown.}
  \label{fig:comp_uobs_heat}
\end{figure*}

Second, we show heatmaps in U-Wall environment in \figref{fig:comp_uobs_heat}.
The results show that \ourmet{} learned more diverse skills than the other methods.

\subsection{\edit{The Number of Discrete Skills in DIAYN}}
\begin{figure*}[tb]
  \begin{subfigmatrix}{4}
 \subfigure[\numocc]{
   \includegraphics[width=0.24\hsize]{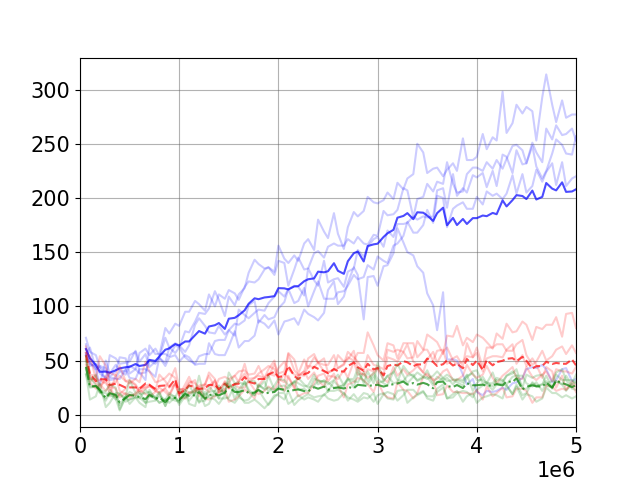}}
 \subfigure[discriminator loss]{
   \includegraphics[width=0.24\hsize]{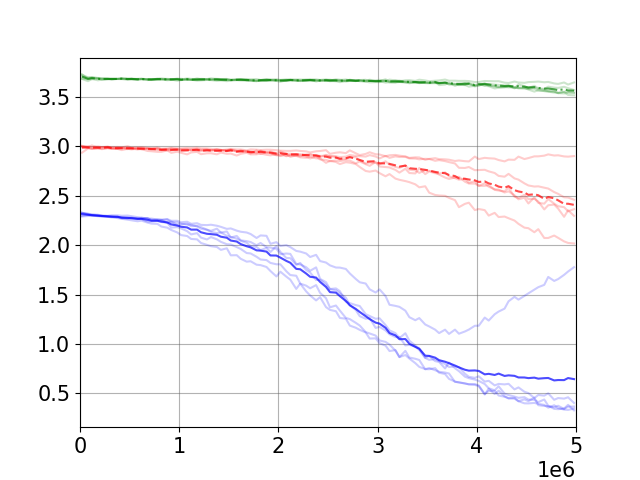}}
 \subfigure[average reward]{
   \includegraphics[width=0.24\hsize]{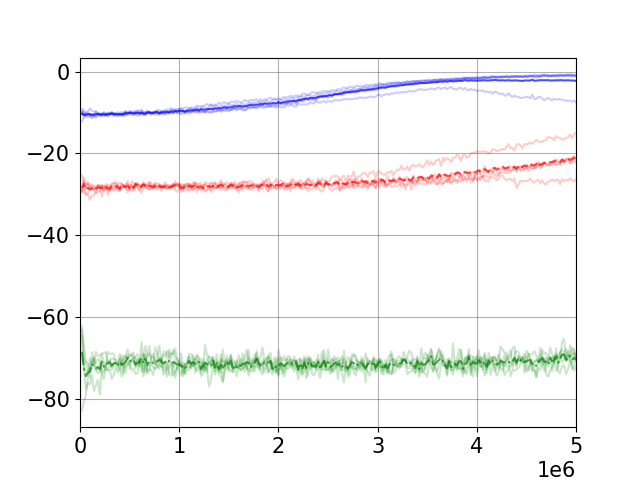}}
 \subfigure[ciritic loss]{                 
   \includegraphics[width=0.24\hsize]{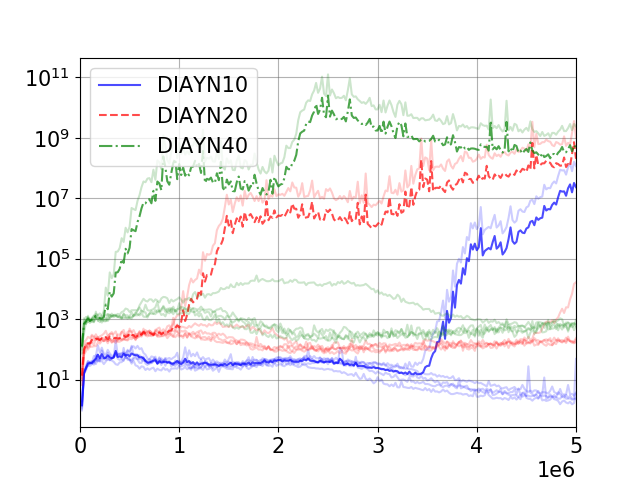}}
 \end{subfigmatrix}
\caption[]{Comparisons of learning curves of DIAYN in NoWall. Data of DIAYN10 and DIAYN40 are the same as those in \figref{fig:comp_visr}}
  \label{fig:comp_diayn}
\end{figure*}

In \secref{sec:difficulties}, we argued that many discrete skills cannot be substitutes for continuous skills.
For more detailed analysis about the argument, we show data of DIAYN whose number of discrete skills is $20$, in addition to the data of $10$ and $40$ in \secref{sec:difficulties}.
The results are shown in \figref{fig:comp_diayn}.
The results of DIAYN20 show intermediate properties and indicate that the sample efficiency decreses as number of discrete skills increases.

\subsection{The Number of Dimensions for Reward Vectors}
\begin{figure*}[tb]
  \begin{subfigmatrix}{4}
 \subfigure[\numocc]{
   \includegraphics[width=0.24\hsize]{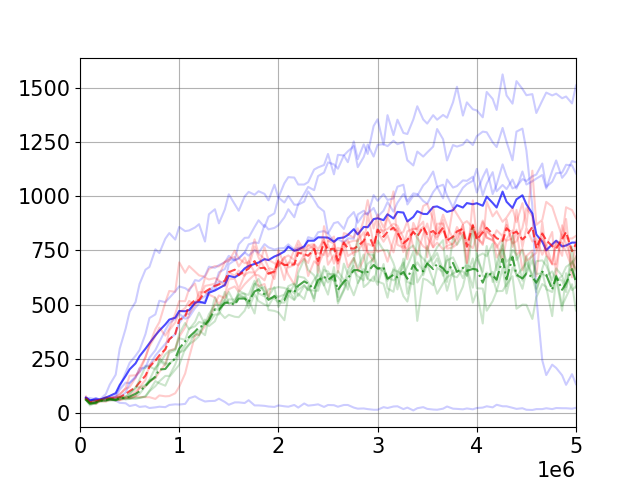}}
 \subfigure[discriminator loss]{
   \includegraphics[width=0.24\hsize]{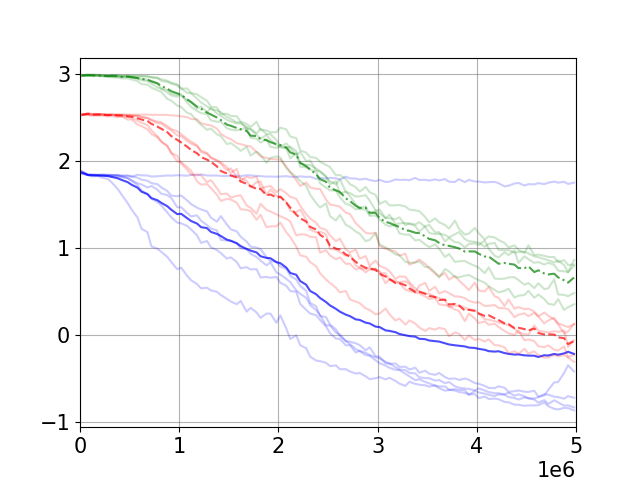}}
 \subfigure[average reward]{
   \includegraphics[width=0.24\hsize]{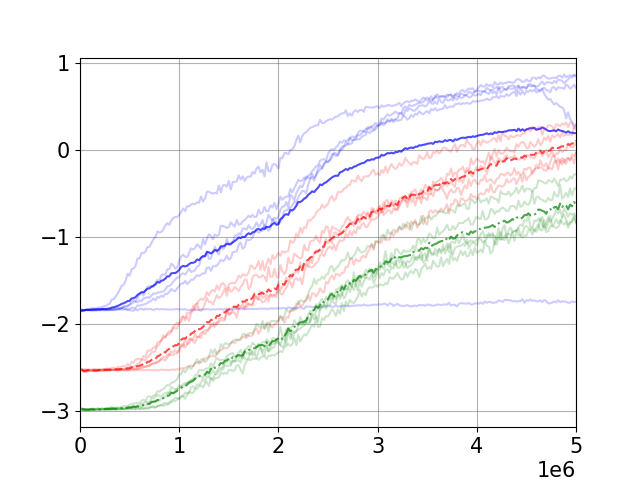}}
 \subfigure[ciritic loss]{                 
   \includegraphics[width=0.24\hsize]{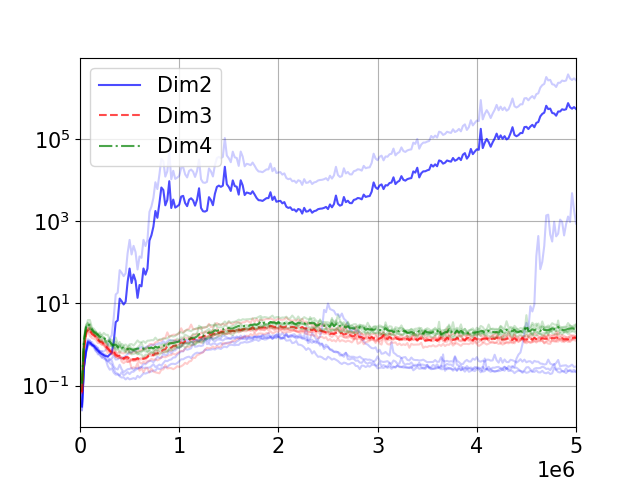}}
 \end{subfigmatrix}
\caption[]{Comparisons of learning curves of NoHIPPS whose number of dimensions of reward vector are 2, 3, and 4 in NoWall. Data of Dim2 are the same as those in \figref{fig:comp_visr}}
  \label{fig:comp_dim}
\end{figure*}

In this section, we examine how much performance of DISCS without HIPPS differs by changing the number of dimensions for the reward vectors.
In \secref{sec:experiments}, the number of dimensions for the reward vectors are two.
In addition to the setting, we examine performance of DISCS without HIPPS when the dimensions are three and four.
The results (\figref{fig:comp_dim}) shows that the average performance worsened as the number of the dimension increased.

\section{Our Implementation Details}\label{sec:implementation}
\begin{algorithm}[tbh]
\caption{\ourmet}
\label{alg:method}
replay buffer $D$, distribution of preference $\P(w)$ 
\begin{algorithmic}[1]
\WHILE{not end}
\STATE $s \leftarrow s_0$ and sample preference $w$ from $\P(w)$ 
\FOR{step in data collection steps}
\STATE Sample action $a$ from policy $\pi(\cdot | s, w)$
\STATE $s' \leftarrow $ env.step($a$)  
\STATE Add \tuple{w, s,a, s'} to $D$ 
\STATE $s \leftarrow s'$
\IF {episode end}
\STATE initilize as line 2 
\ENDIF
\ENDFOR
\IF {update discriminator}
\STATE Sample \tuple{w, s, a, s'} from $D$
\STATE Update $q_\phi$
\ENDIF
\FOR{step in update steps}
\STATE Sample \tuple{w, s, a, s'} from $D$ and sample hindsight preferences $w'$ from $q_\phi(w|s)$ 
\STATE Generate reward vector $r$ and $r'$ for \tuple{w, s, a, s'} and \tuple{w', s, a, s'} 
\STATE Update $Q$ and $\pi$ by using \tuple{w, s, a, r, s'} and \tuple{w', s, a, r', s'} 
\STATE Update Q-target 
\ENDFOR
\ENDWHILE
\end{algorithmic}
\end{algorithm}
We implemented our method by modifying SAC in stable-baseline3~\cite{stable-baselines3}.
A summary of the modification from the SAC implementation is provided here.
\begin{itemize}
\item Change SAC to MOSAC\\
  More concretely, we implemented the preference conditional policy and Q-network. 
  Also, we modified the replay buffer to preserve preferences in rollouts.
\item Implement discriminator $q_\phi(s_t|w)$ and its training procedure
\item Implement the partial derivative of $\log C_m(\kappa)$ with regard to $\kappa$\\
  We used the modified Bessel function in SciPy.
  To backpropagate the gradients, their calculation has to be implemented manually. 
\item Implement HIPPS
\end{itemize}
We will release the code when the paper is accepted.

A summary of hyperparameters in our experiments is provided as \tbref{tb:hyperparams}.
Although this is ommitted for simplicity of explanation in \algref{alg:method}, $\pi$ and $\bar{\theta_Q}$ is not updated every loop.
For the sake of clarity, we show the number of updates of each network per timesteps.
Incidentally, ``data collection steps'' and ``update steps'' in the pseudo code are both 8.
\begin{table*}[hbtp]
  \centering
  \begin{tabular}{ll}
    \hline
    \edit{the number of dimensions in reward vectors}   & 2\\
    the number of Q-networks   & 2\\
    size of hidden layers in Q-networks   & 256, 256, 64\\
    size of hidden layers in policy network   & 256, 256 \\                                             
    size of hidden layers in discriminator   & 256, 256 \\                                             
    the number of Q updates per timesteps & 1                                \\
    the number of Q-target updates per timesteps& 1/8                                \\
    the number of policy updates per timesteps  & 1/8 \\
    discriminator update per timesteps & 1/50000 \\
    batch size  & 1024 \\
    batch size for discriminator updates & 16384  \\
    replay buffer size & 2e+6  \\
    $\gamma$ & 0.99  \\
    $\alpha$, i.e., entropy coeffient   &  0.1 \\
    $\tau$, i.e., learning rate of Q-target   &  0.005 \\
    optimizer    &  Adam \\
    learning rate    &  3e-4 \\
    \hline
  \end{tabular}
  \caption{The hyperparameters in our experiments.}
  \label{tb:hyperparams}
\end{table*}

Although there is one critic in explanation in \secref{sec:mosac} for simplicity, in our actual implementation, two Q-functions are used in the same way as SAC.
In this case, the parameter vectors of Q-targets are updated as follows
\begin{align}
  \bar{\theta}_i \leftarrow \tau \theta_i + (1- \tau)\bar{\theta}_i, \; (i =1, 2),
  \label{eq:target_update2}
\end{align}
and Q-targets are calculated as follows
\begin{align}
  Q_{\bar{\theta}_Q} = \arg\min_{Q \in \{Q_{\bar{\theta}_1}, Q_{\bar{\theta}_2}\}} \tilde{w}^\top Q.
  \label{eq:q_target2}
\end{align}
We use the following critic loss to train each Q-network:
\begin{align}
&\E\left[- (\tilde{w}^\top(Q_{{\theta_{Q}}_i}(s_t,a_t,w) - \hat{\T}Q_{\bar{\theta}}(s_t,a_t,w)))^2 \right]\; (i =1, 2), \;\mbox{where}\label{eq:critic_loss2}\\
&\hat{\T}Q_{\bar{\theta}}(s_t,a_t,w) = \tilde{r}_\phi(s_t) + \gamma \E_{a_{t+1} \sim \pi(\cdot|s_{t+1}, w)}[Q_{\bar{\theta}}(s_{t+1},a_{t+1},w) + h^{\pi}(s_{t+1},a_{t+1},w)].
\end{align}

\end{document}